\documentclass{article}

\usepackage{PRIMEarxiv}

\usepackage[utf8]{inputenc} 
\usepackage[T1]{fontenc}    
\usepackage{hyperref}       
\usepackage{url}            
\usepackage{booktabs}       
\usepackage{amsfonts}       
\usepackage{nicefrac}       
\usepackage{microtype}      
\usepackage{lipsum}
\usepackage{graphicx}
\graphicspath{{media/}}     
\usepackage{mathtools}
\usepackage[toc,page]{appendix}
\usepackage{subcaption}
\usepackage{mwe}
\usepackage{mhchem}
\usepackage{csquotes}
\usepackage{siunitx}
\usepackage[casechanger=auto,style=ext-numeric-comp, sorting=none, giveninits=true, isbn=false, doi=false,date=year,minbibnames=4, maxbibnames=4, articlein=false]{biblatex}
\DeclareFieldFormat[article,inbook,incollection,inproceedings,patent,unpublished]{titlecase:title}{\MakeSentenceCase*{#1}}

\title{Materium: An Autoregressive Approach for Material Generation
}

\author{
  Niklas Dobberstein, Jan Hamaekers \\
  Fraunhofer Institute for Algorithms and Scientific Computing SCAI \\ Schloss Birlinghoven \\ 53757 Sankt Augustin, Germany.\\
  \texttt{\{niklas.dobberstein, jan.hamaekers\}@scai.fraunhofer.de} \\
}

\begin{document}
\maketitle

\begin{abstract}
We present Materium: an autoregressive transformer for generating crystal structures that converts 3D material representations into token sequences. These sequences include elements with oxidation states, fractional coordinates and lattice parameters. Unlike diffusion approaches, which refine atomic positions iteratively through many denoising steps, Materium places atoms at precise fractional coordinates, enabling fast, scalable generation.
With this design, the model can be trained in a few hours on a single GPU and generate samples much faster on GPUs and CPUs than diffusion-based approaches.
The model was trained and evaluated using multiple properties as conditions, including fundamental properties, such as density and space group, as well as more practical targets, such as band gap and magnetic density. In both single and combined conditions, the model performs consistently well, producing candidates that align with the requested inputs.
\end{abstract}


\section{Introduction}

Advances in materials science have had a significant impact on many different industries, including energy, electronics, transportation and manufacturing.  Therefore, discovering and designing materials with specific properties is
of high relevance in practical applications. However, exploring the vast combinatorial chemical space remains a significant challenge. Traditional discovery methods rely on laborious experimental workflows, which are resource-intensive and slow to scale. To address this, much of the effort has shifted to \textit{in silico} physics-based methods such as Density Functional Theory (DFT) \cite{PhysRev.140.A1133,PhysRev.136.B864, Argaman_2000}. While these simulations have enabled the systematic screening of materials, their high computational cost per candidate is still a huge bottleneck for large datasets.

Machine learning (ML) methods, and deep learning (DL) in particular, have become increasingly popular over the last decade. 
Generative models have emerged as a central tool in these fields. Prominent examples include diffusion models \cite{ho2020denoisingdiffusionprobabilisticmodels, score_based_gen_model}, flow matching (FM) \cite{lipman2023flowmatchinggenerativemodeling}, and autoregressive language models \cite{radford2019language, attention}. 
For the generation of bulk materials, diffusion-based approaches such as CDVAE~\cite{cdvae_xie_grossman_2018}, UniMat~\cite{unimat_paper} and MatterGen~\cite{mattergen_nature} have been developed. Additionally, flow matching methods operating with periodic boundary conditions, such as FlowMM~\cite{miller2024flowmmgeneratingmaterialsriemannian}, have been introduced to better respect crystal symmetry and periodicity.
Finally, language model-based methods have been presented, such as LLaMat~ \cite{mishra2025foundationallargelanguagemodels} and CrystaLLM~ \cite{antunes2024crystalstructuregenerationautoregressive}, which generate crystal structures using the Crystallographic Information File (CIF) format. However, the latter approaches are not very efficient in terms of token use, as single coordinates are split into multiple character tokens (e.g. `0', `.' `1', `2', `3'), resulting in long, inefficient sequences. 
An alternative is to use the Simplified Line-Input Crystal-Encoding System (SLICES)~\cite{Xiao2023, chen2024mattergptgenerativetransformermultiproperty}, which provides an invariant representation of a crystal's topology. However, the SLICES representation has the drawback that the model learns an abstract representation of the material that must that must be converted back to the correct crystal structure via a complex pipeline based on ML models like M3GNet~\cite{Chen2022}, which can introduce additional inaccuracies.


In this paper we introduce Materium, an autoregressive language modelling approach for material generation inspired by Meshtron \cite{hao2024meshtronhighfidelityartistlike3d}. 
Here, we tokenize the materials components, fractional coordinates, atomic elements and lattice directly, enabling the model to generate very precise 3D structures. 
This is done by discretizing the continuous values into 1024 tokens, which will be discussed more in the later sections.
In comparison with recent state-of-the-art generators (e.g., MatterGen), Materium has a comparable parameter count of 43 million parameters, yet trains substantially faster, reaching convergence in about 4 hours on a single A100 GPU with 600k materials compared to multiple days for MatterGen.
Finally, we include capabilities for conditional generation of materials with specific target properties such as band gap and magnetic density.
 
The remainder of this paper is organized as follows. In Section~\ref{sec:tokenization}, we describe our method for representing crystalline materials for generative modeling. In Section \ref{sec:architecture}, we present the model architecture and in Section \ref{sec:training} we provide details about the data curation and the model training. In Section \ref{sec:evaluation} we discuss the evaluation of our model in terms of sequence order sensitivity and performance for unconditional and conditional generation, before we conclude our findings in Section~\ref{sec:conclusion}.

\section{Tokenization of Crystal Structures}
\label{sec:tokenization}

This section details our method for representing crystalline materials for generative modeling. We first define the fundamental components of a crystal structure and then describe how these components are converted into an ordered, one-dimensional sequence of tokens for our autoregressive model. 

\subsection{Crystal Definition}

A crystal structure is defined by three fundamental components. 
The first of these are the three lattice vectors (written as a $L \in \mathbb{R}^{3 \times 3}$ matrix), which describe the the shape and orientation of the periodic cell.
In our case, this will be the smallest possible periodic cell, or unit cell. 
Note that such a unit cell has just six degrees of freedom. Commonly used so-called cell or lattice parameters are the lengths of the cell edges, namely $a$, $b$ and $c$, and the angles between them, namely $\alpha$, $\beta$ and $\gamma$.
The atoms inside the cell can then be split into position and element components. It should be noted that the complete crystal structure would then be this unit cell repeated periodically in all three ($x, y, z$) directions.

The positional component describes the three dimensional ($x, y, z$) position of each atom in the material, however there is a distinction in how we can represent them. For $N$ atoms, one version is to use the Cartesian coordinates ($\mathbf{r}_{\text{cart}} \in \mathbb{R}^{3N}$), which are coordinates in {\em physical} space.
Alternatively, since a unit cell obey periodic boundary conditions, one can use fractional coordinates, which we opted for in our model. Fractional coordinates, $\mathbf{r}_{\text{frac}} \in [0..1)^{3N}$, are relative positions expressed in the basis of the lattice vectors ($\mathbf{l}_1, \mathbf{l}_2, \mathbf{l}_3$). The conversion to Cartesian coordinates is given by the lattice matrix $\mathbf{L} = [\mathbf{l}_1, \mathbf{l}_2, \mathbf{l}_3]$:
\[
\mathbf{r}_{\text{cart}} = \mathbf{L} \cdot \mathbf{r}_{\text{frac}}
\]
Each component of a fractional coordinate vector is confined to the interval $[0, 1)$, which means the atomic positions are defined within the unit cell on a 3D torus. We choose to use fractional coordinates because their bounded range makes it much simpler to quantize them into a discrete set of tokens, whereas the unbounded nature of Cartesian coordinates complicates this process.
Lastly, attached to the position of the atom we also need the atomic element, i.e. oxygen (O), hydrogen (H) etc. 

However, simply knowing the atomic element is not always sufficient to describe the chemical state of the atom within the crystal. 
For this reason, we also incorporate the oxidation state. 
The oxidation state of an atom represents the number of electrons it has lost when forming chemical bonds with its neighbors. 
For instance, iron (Fe) can commonly exist in a $+2$ or $+3$ state, while oxygen (O) is typically found in a $-2$ state. A stable crystal structure should have a net charge of zero, meaning the sum of all positive and negative oxidation states of its constituent atoms must balance out.

\subsection{Tokenization Process}

A fundamental challenge in applying autoregressive models to material generation is the need to convert an inherently unordered set of atoms into an ordered, one-dimensional sequence of tokens. However, the choice of this ordering can significantly influence the model's ability to learn the underlying patterns of crystal structures. To address this, we explored several deterministic and stochastic ordering strategies for the atomic sites within a crystal.

Our approach begins by representing a crystal structure as a sequence of discrete tokens. This sequence is composed of special tokens to denote the start (\texttt{[SOS]}), end (\texttt{[EOS]}), and different sections of the structure (\texttt{[ATOMS]}, \texttt{[LATTICE]}), as well as tokens for each atomic element and the quantized values of fractional coordinates and lattice parameters.
Additionally, for each atomic element we include their common oxidization states plus the neutral oxidization state $0$, which are taken from pymatgen's  \cite{pymatgen} heuristics.
This means that now each atomic element and oxidization state pair is a token in our vocabulary. 
For example, they are represented like this \texttt{[O|-2]}, \texttt{[Fe|+3]}, \texttt{[O|0]}.
Using this technique it should be easier for the model to keep track of the charges and create neutrally charged materials.
Before tokenization, each crystal structure is standardized by reducing it to its Niggli primitive cell\footnote{A Nigli primitive cell is a specialized type of primitive cell used in crystallography to provide a unique and standardized mathematical description of a crystal lattice~\cite{grosse2004numerically}.} to ensure a canonical representation using pymatgen.

\begin{figure}[ht!]
    \centering
    \includegraphics[width=0.8\textwidth]{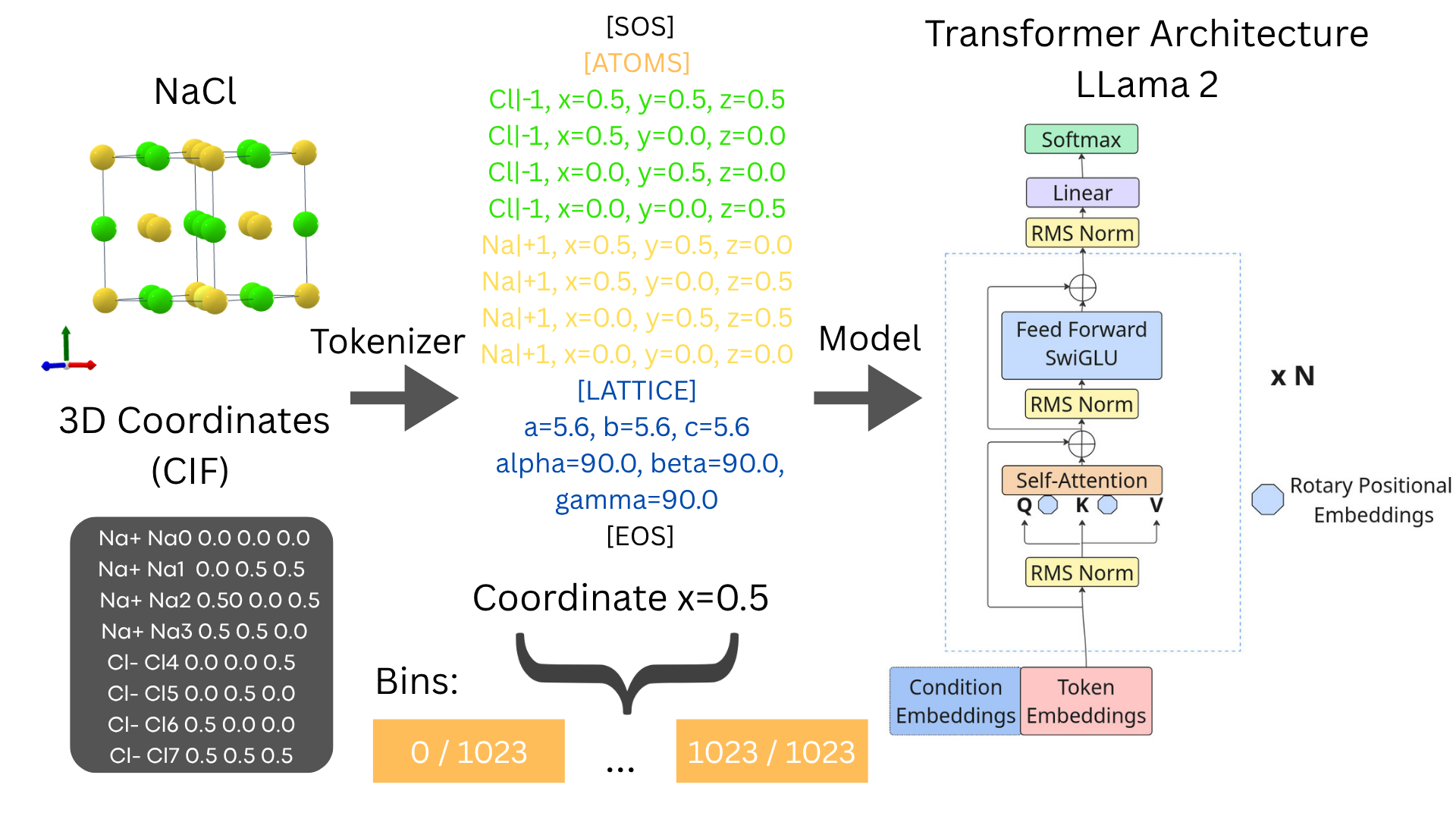}
    \caption{Visualization of the tokenizer and model architecture used in Materium.}
    \label{fig:model_vis}
\end{figure}

The final tokenized representation of a crystal follows a structured format. The sequence always begins with a \texttt{[SOS]} token, followed by the \texttt{[ATOMS]} section. This section is composed of a sequence of atoms, where each atom is tokenized into a quadruplet of four tokens: one token for the atomic element type, and three tokens for the quantized x, y, and z fractional coordinates. Following the atoms, the \texttt{[LATTICE]} section contains six tokens representing the lattice parameters, and the entire sequence concludes with an \texttt{[EOS]} token. A complete sequence therefore looks like:
\texttt{[SOS] [ATOMS] [atom1\_element|oxi\_state] [atom1\_x] [atom1\_y] [atom1\_z] [atom2\_element|oxi\_state] ... [atom\_N\_z] [LATTICE] [a] [b] [c] [alpha] [beta] [gamma] [EOS]}.
We also tested a model by changing the internal order of the atoms by placing the positional tokens, i.e. [atom1\_x] [atom1\_y] [atom1\_z],  first and then the atom type, i.e. [atom1\_element|oxi\_state], but we could not overserve a significant impact on the performance of the model. An illustration of the process can be seen in the Figure \ref{fig:model_vis}.

The tokenization process involves two main components: the arrangement of atoms and the discretization of continuous parameters. For the atomic arrangement, we investigated four distinct ordering schemes. The first one sorts atoms based on their atomic number in ascending order, with a secondary sort based on their fractional coordinates, namely $x$, $y$, then $z$. The second approach reverses this, sorting by the heaviest elements first. The third method prioritizes the spatial location, sorting first by the positional coordinates and then by the atomic number. As a last approach we just use random order for the atoms to have a baseline to compare to if the sequence order has a large impact.  

To handle the continuous nature of fractional coordinates and lattice parameters, we employed a quantization technique. Each continuous value is discretized into one of 1024 bins. For fractional coordinates, the values are inherently within the [0, 1) range so each token has a resolution of $\frac{1}{1024}$. The $1024$-tokens are shared across the x,y and z direction
For the six lattice parameters (a, b, c, $\alpha$, $\beta$, $\gamma$), we first normalize them based on fixed ranges determined from our training dataset before quantization. The ranges used are $a \in [2.0, 10.0]$ \text{\AA}, $b \in [2.0, 12.5]$ \text{\AA} , $c \in [2.0, 20.0]$ \text{\AA}, and $\alpha, \beta, \gamma \in [60.0, 120.0]$ degrees.
Using this approach we can effectively create a sequence from any material structure and use standard language modeling techniques to train on it.

\section{Model Architecture} \label{sec:architecture}

Materium is a decoder-only transformer designed for autoregressive generation, with an architecture based on the Llama 2 architecture~\cite{touvron2023llama2openfoundation} as shown in the Figure \ref{fig:model_vis}. The model is  small, with approximately 43 million parameters, and is composed of 12 decoder blocks.
In total, the model uses 16 attention heads per attention layer and a hidden dimension of 512. The feed forward network uses 1536 parameters for the hidden dimension. For regularization we use dropout layer \cite{JMLR:v15:srivastava14a} with 10\% dropout rate.
Because of the small size of the model even the generation on CPU is quite fast.

Each decoder block follows a pre-normalization structure, beginning with an RMSNorm layer \cite{zhang2019rootmeansquarelayer}. This is followed by a masked multi-head self-attention layer \cite{attention} and a subsequent feed-forward network (FFN). Another RMSNorm layer is applied before the FFN. For the FFN, we employ the SwiGLU activation function \cite{shazeer2020gluvariantsimprovetransformer}, which has been shown to improve performance over standard ReLU variants. Given the model's modest scale, we utilize a full multi-head attention mechanism rather than more complex variants like Grouped-Query Attention (GQA) \cite{ainslie2023gqa}.

The masked multi-head self-attention layer allows the model to weigh the importance of different tokens in the input sequence when predicting the next token. For an input sequence embedding $X \in \mathbb{R}^{L \times d_{emb}}$, the attention mechanism is defined as:
\begin{align*}
\text{MMHA}(X) &= \text{Concat}(\text{head}_{1}, \ldots , \text{head}_{h}) \cdot W_O \\
\text{head}_i &= \text{MaskedAttention}(X W_Q^i, X W_K^i, X W_V^i) \\
\text{MaskedAttention}(Q, K, V) &= \text{softmax} \left(\frac{QK^T}{\sqrt{d_k}} + M \right) V
\end{align*}
Here, $Q$, $K$, $V$ are the query, key, and value matrices, respectively, generated through linear transformations, $W_Q$, $W_K$, $W_V$ respectively, of the input $X$. The causal mask $M$ is a lower-triangular matrix of zeros and negative infinities, ensuring that the prediction for a token at position $i$ can only depend on the preceding tokens. $d_k$ is the hidden-dimension of the K matrix and used as a normalization in the softmax. 
We employ Rotary Positional Embeddings (RoPE) \cite{su2023roformerenhancedtransformerrotary} over the entire sequence to remove the permutation equivariant properties of the Transformer. 

To enable controlled generation, the model can be conditioned on various material properties. Each numerical property is transformed into a value embedding via two embeddings. First, the scalar value of a property is projected into the embedding dimension $d_{emb}$ of the model using a property-specific linear layer to create a value embedding.
Additionally, an embedding vector is added to the value embedding for each property label, which is learnt by the model so that it can differentiate between different properties, even if the order changes. 
The reduced formula is treated slightly differently, as it is first decomposed into atomic elements and their respective stoichiometry. Each of these is embedded separately via its own lookup table. The resulting atomic element and stoichiometry embeddings are then added together again. 

A significant challenge in materials datasets is that not all properties are available for every material. 
To make the model robust to such missing information and to prevent it from becoming dependent on a fixed set of conditions, we employ a conditional dropout strategy during training \cite{Dobberstein2024}.  
For each training batch, we randomly omit each condition with a 50\% probability.
This technique forces the model to learn to generate plausible structures from any subset of the provided conditions, including none at all. 
However, during a batch some conditions might still be not given, to enable efficient batched training we therefore replace the NaN values with a learned NaN-embedding.
The final conditional embeddings are prepended to the main material token sequence.

\section{Training}
\label{sec:training}


\subsection{Dataset}

To train our generative model, we utilize the large and diverse dataset of crystalline materials presented in the MatterGen paper \cite{mattergen_nature}, which is referred to as Alex-MP-20. 
This dataset was specifically curated for training generative models of inorganic materials and is built upon a comprehensive reference set of approximately \num{1.1} million structures, known as Alex-MP-ICSD.

The reference set combines structures from major materials databases, primarily the Materials Project (MP) \cite{Jain2013} and the Alexandria database (Alex) \cite{Schmidt2024}, with additional data from the Inorganic Crystal Structure Database (ICSD) \cite{nist_icsd} in the Alex-MP-ICSD reference set. 
All structures in the Alex-MP-ICSD dataset were recalculated using DFT to be consistent with MP. 
The Alex-MP-20 only includes the materials from the Alex-MP-ICSD dataset with energy above the convex hull of less than \SI{0.1}{\eV\per\text{atom}}, which is a measure of thermodynamic stability \cite{Aykol2018} and having less than 20 atoms in the unit cell.
In total, the dataset has 675,204 materials with various associated properties.

The model was designed to generate materials conditioned on a diverse set of target properties. 
Therefore we used a wide range of conditions provided by the dataset. These conditions span several categories, including fundamental physical properties such as the \textit{band gap} and \textit{magnetic density}.
We also include intrinsic structural properties like the material's \textit{density} and \textit{space group}, as well as the chemical \textit{reduced formula}. 
Furthermore, to explore conditioning on properties beyond the purely physical, we incorporate the \textit{Herfindahl-Hirschman Index (HHI)}, a socio-economic metric that quantifies elemental scarcity and supply concentration \cite{Gaultois2013}.

We did additional small preprocessing steps for the conditions to have them all in the same numerical range of about $[-10,10]$. 
For properties which have values near zero, such as the band gap (in \si{\eV}) and the magnetic density (in \si{\per\cubic\angstrom}) we applied a logarithmic transformation of the form $\log(x + 10^{-3})$. This was done to improve the numerical stability and resolution in the dense $[0.0, 0.2]$ range. 
Additionally, the HHI was divided by 1000 to \enquote{normalize} it to the same range as the other values.

\subsection{Training Procedure}

We trained the model on a single A100 40GB VRAM GPU in about 4 hours for 50 epochs using PyTorch \cite{paszke2019pytorchimperativestylehighperformance} and FlashAttention \cite{dao2022flashattentionfastmemoryefficientexact}. Additionally, we used 16-bit brain floating point (bfloat16) for more efficient training. 
In total, we used a batch size of $512$ with the AdamW optimizer \cite{loshchilov2019decoupledweightdecayregularization} using a starting learning rate of \num{4e-4} which is reduced by $50\%$ if after $3$ epochs the validation loss did not decrease. 
The dataset Alex-MP-20 was split into a training dataset of size $607,683$ ($90\%$) and a test set of size $67,521$ ($10\%$).
The model is trained using cross-entropy loss to predict the next token in the sequence. However, this only applies to the token sequence of the materials and not to the prepended conditioning.

Here, the tokenised material is first prepended with the given conditioning information, embedded as a $d_{emb}$-dimensional vector, and then passed through the Llama 2 architecture. 
Finally, the model predicts the probability distribution of the next-token in an autoregressive manner \cite{radford2019language,brown2020languagemodelsfewshotlearners}.

\section{Evaluation}
\label{sec:evaluation}

In this section, we thoroughly evaluate the model in terms of sequence order sensitivity and performance for unconditional and conditional generation. 
In addition, we tested the model on multi conditional generation as this is already integrated in the training procedure. 
However, we can not reasonably test all combinations as we trained on five conditions. 
Therefore we selected a couple of interesting and practical combinations to evaluate. 

For all generations we used a temperature of $1.0$ as this gave us the most diversity with good stability for the generated materials. 

\subsection{Sensitivity to Token Ordering and Unconditional Generation}
Since an ordered sequence has to be created from an inherently unordered object, many different possibilities exist. In this work, we explored four simple strategies:
\begin{enumerate}
    \item Sort by increasing atomic number first, i.e. \ce{H, He, Li}, \dots and then sort after $x$, $y$ and lastly $z$ fractional coordinates.
    \item Use the highest atomic number first, i.e. \dots \ce{Li, He, H} and then sort after $x$, $y$ and lastly $z$ fractional coordinates.
    \item Sort using the fractional coordinates, i.e.\, first $x$, then $y$ and then $z$. \footnote{This provides a deterministic ordering, as no two atoms share the same coordinates.} 
    \item Use a random ordering of the elements as a baseline, although the relative ordering of the $x$, $y$, $z$ coordinates of course stays the same.
\end{enumerate}

For the random ordering we only used one random permutation per material to speed up the training time.

To investigate the impact of token ordering, we trained four distinct models, each employing one of the strategies above, but otherwise having the same amount of trainable parameters and architecture. 
We refer to these models $M_{low}$ for the smallest atomic number first ordering, $M_{high}$ for the largest atomic number first, $M_{xyz}$ for the coordinate-based ordering, and $M_{rand}$ as the random permutation baseline.

We first evaluated the models on their unconditional generation capabilities. For each model, we generated a set of $1024$ crystal structures using a sampling temperature of $1.0$. The quality of these generated materials was assessed using the evaluation methods described in the MatterGen paper \cite{mattergen_nature} and use the Github implementation, which provides metrics for stability, uniqueness, and novelty. 
We used MatterSim-v1.0.0-5M inter-atomic potential model \cite{yang2024mattersimdeeplearningatomistic} to relax the materials as using DFT would have been to computationally expensive and time-consuming. 
However, we applied DFT using the Quantum Espresso (v. 7.4.1) software \cite{Giannozzi_2009}, to calculate and compare the band gap in the conditional generation case.
Stability was determined by the energy above the convex hull, with a lower value indicating a more stable structure. Uniqueness measures the diversity of the generated samples, while novelty quantifies the percentage of generated structures not present in the training data \cite{mattergen_nature}. 

\begin{table}[h]
\centering
\caption{Unconditional generation performance on 1024 generated and relaxed materials using MatterSim for different token ordering strategies. Lower energy above hull, lower RMSD, and a higher fraction of stable, unique, and novel structures are better.}
\label{tab:unconditional_results}
\begin{tabular}{lccc}
\hline
\textbf{Model} & \textbf{Avg. Energy Above Hull} (\si{\eV\per\text{atom}}) & \textbf{Frac. Stable, Unique, Novel} (\%) & \textbf{Mean RMSD} (\si{\angstrom}) \\ \hline
$M_{low}$    & 0.098                                     & 20.9 & 0.208 \\
$M_{high}$    & 0.097                                     & 22.7 & 0.208 \\
$M_{xyz}$      & 0.082                                     & 22.0 & 0.142 \\
$M_{rand}$     & 0.164                                     & 18.7 & 0.448 \\ \hline
\end{tabular}
\end{table}
The results from our unconditional generation experiments, summarized in Table \ref{tab:unconditional_results}, revealed a trade-off between the different ordering strategies. 
The $M_{xyz}$ model demonstrated superior performance, compared to the other models in the table, in generating highly stable structures, achieving both the lowest average energy above the convex hull and the lowest mean RMSD. This suggests that a coordinate-based ordering helped the model generate structures that are already close to their relaxed, low-energy state.

The $M_{high}$ model had the highest stability, uniqueness and novelty (S.U.N) rate and a very low average energy above hull. 
However, the S.U.N rate of $22.7\%$ is very low compared to state of the art models like MatterGen, which reports a S.U.N rate of $38.9 \%$. Additionally, our model performed slightly worse than MatterGen in the average energy above hull $0.09$ and their average RMSD $0.06$.

We investigated the components of the S.U.N rating and saw that our model had a very high uniqueness ($99.8\%$) percentage and a good stability ($73.9\%$) percentage. However, the percentage of novel structures ($46.5\%$) generated was relatively low.
To counteract this we tried increasing the temperature and tried different temperatures for the different token types, i.e. higher temperatures for the atom-types and/or coordinates. However, these changes never helped to improve the S.U.N score, because when we generated more novel materials the stability score decreased almost the same amount or even more.

The major advantage of our model is the training and generation speed compared to the diffusion models. As previously mentioned, a training run took around four hours on a single A100 40 GB VRAM GPU with a time of around six minutes per epoch. We used the provided MatterGen code to train on the same GPU and it takes around three days with an average of 42 minutes per epoch.

Additionally, we observed that our model can generate materials much quicker than MatterGen using version mattergen\_base with default settings. On an Intel Xeon Gold 6130 CPU using 8 cores out of the 32 a generation of a single material took MatterGen three minutes, while ours only took around $1.5$ seconds. Even on a single A100 GPU our model is much faster as it can generate a material in less than a second compared to about one minute for MatterGen.

As we show in the following sections, our model also shows the most consistent performance across various conditional generation tasks. Therefore, we selected the $M_{high}$ model, which we will also refer to as the Materium, for all subsequent evaluations in this study.

\subsection{Unconditional Generation $§M_{high}$}

We further analysed the unconditional generation results of the $M_{high}$ model. We compared the match rate, which measures how many of the generated structures could be successfully matched to their relaxed counterparts. This matching was performed using \texttt{pymatgen}'s \texttt{StructureMatcher} module, which compares crystal structures by accounting for periodicity and allowing for different lattice settings and atomic permutations. Out of $1024$ generated samples, we found a successful match for 874, resulting in a high match rate of $85.3\%$. 
This high match rate strongly indicates that the model can generate materials near their local energy minima successfully and 
it is also a very good match rate compared to other material generation models such as FlowMM \cite{miller2024flowmmgeneratingmaterialsriemannian} ($61.39 \%$), DiffCSP \cite{jiao2024crystalstructurepredictionjoint} ($51.49 \%$) and CDVAE \cite{xie2022crystaldiffusionvariationalautoencoder} ($45.31$) which are trained on the MP20 dataset. However, in their cases, they  only used the MP20 dataset, not including the Alexandria dataset to train, so the are not directly comparable, but it gives a good validation that our approach works.

Furthermore, we examined the distribution of the RMSD in Figure \ref{fig:rmsd_dist} between the generated and relaxed materials by relaxing them using MatterSim. There we observed that most of the generated materials have a really low RMSD, meaning that the materials before the relaxation are very similar to the ones after the relaxation. Specifically, $698 / 887 = 78.7 \%$ of the matching materials were below \SI{0.1}{\angstrom} for the RMSD. This demonstrates that the model can generate material structures that are very close to their energetic minimum.  

While these experiments only used MatterSim and not DFT these results are not directly comparable to, for example, MatterGen, as all of their experiments were validated using DFT as it is more robust and accurate than machine-learning models. 

\subsection{Single-condition Comparison between Models}

Following the unconditional evaluation, we assessed the models on several, easy to evaluate, single-conditional generation tasks to understand how token ordering affects performance on more constrained problems. We focused on the generation of materials conditioned on a target space group, reduced formula, and density as they are the easiest and fastest to validate. However, for this conditional task we left out the $M_{rand}$ model as it performed much worse than the models which had an order built in.

\begin{figure*}[t]
  \centering

  \begin{subfigure}{0.48\textwidth}
    \centering
    \includegraphics[width=\linewidth]{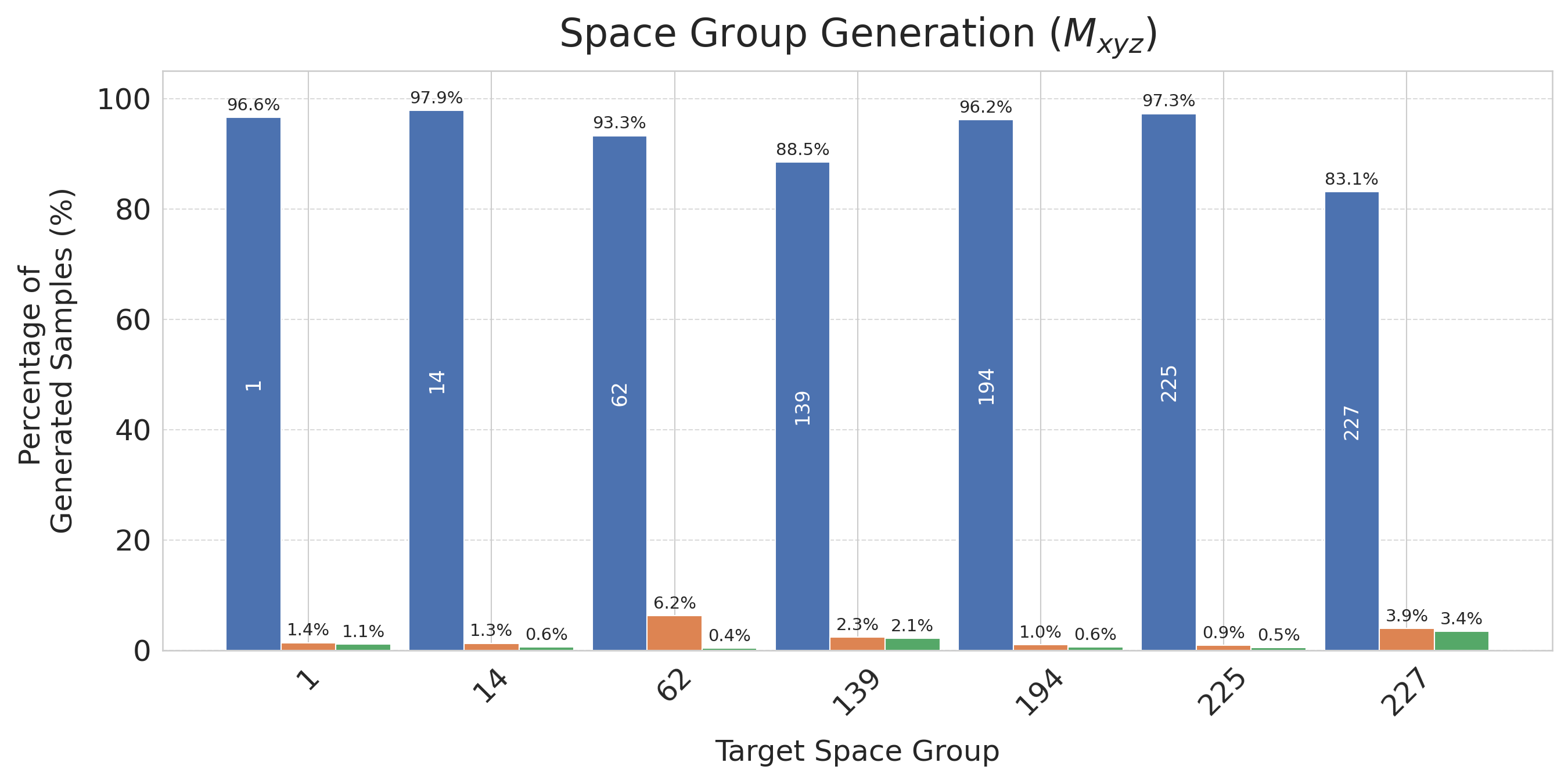}
    \caption{Space group — $M_{xyz}$}
    \label{fig:sg_xyz}
  \end{subfigure}\hfill
  \begin{subfigure}{0.48\textwidth}
    \centering
    \includegraphics[width=\linewidth]{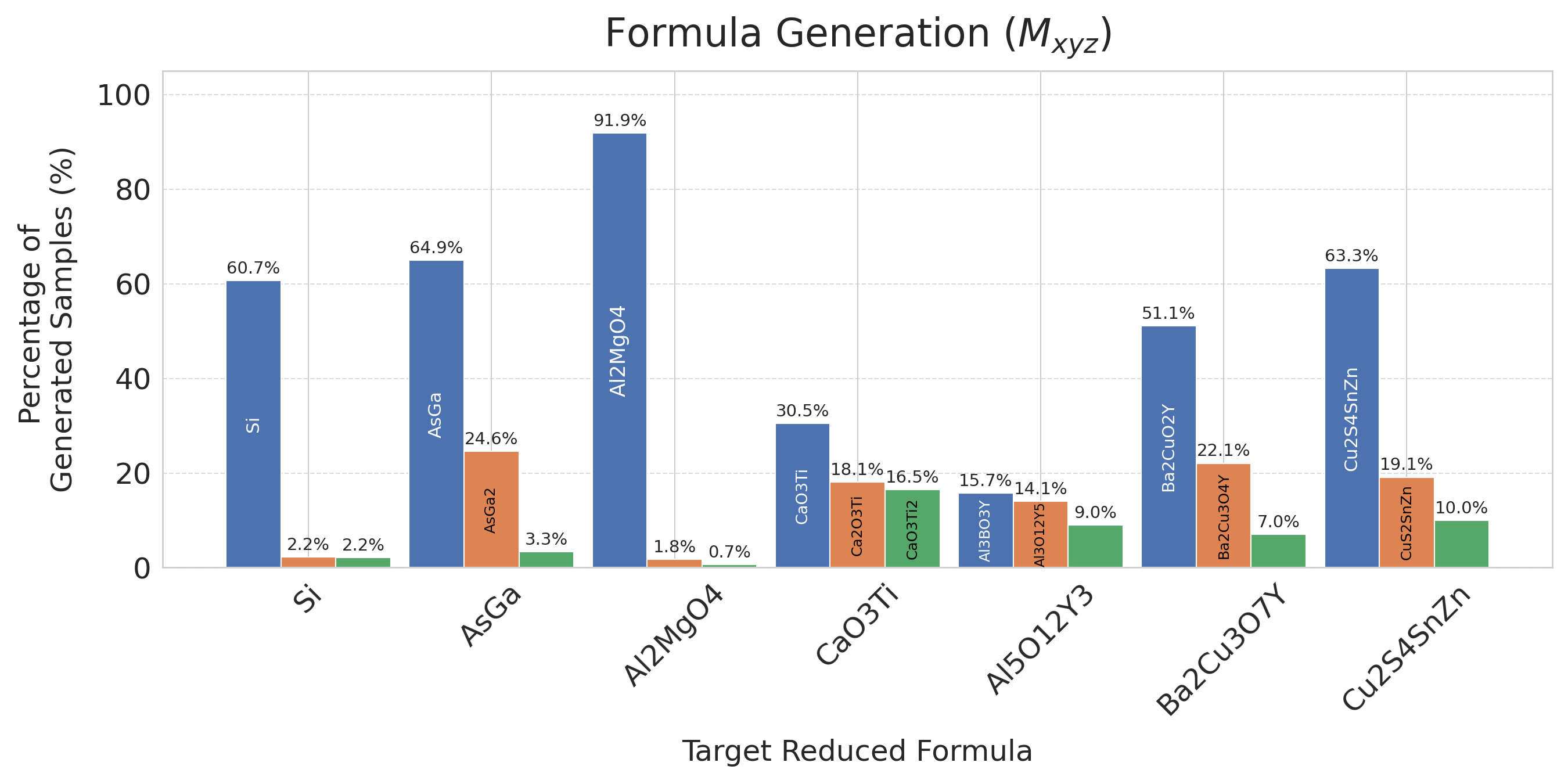}
    \caption{Formula — $M_{xyz}$}
    \label{fig:formula_xyz}
  \end{subfigure}

  \vspace{0.6em}

  \begin{subfigure}{0.48\textwidth}
    \centering
    \includegraphics[width=\linewidth]{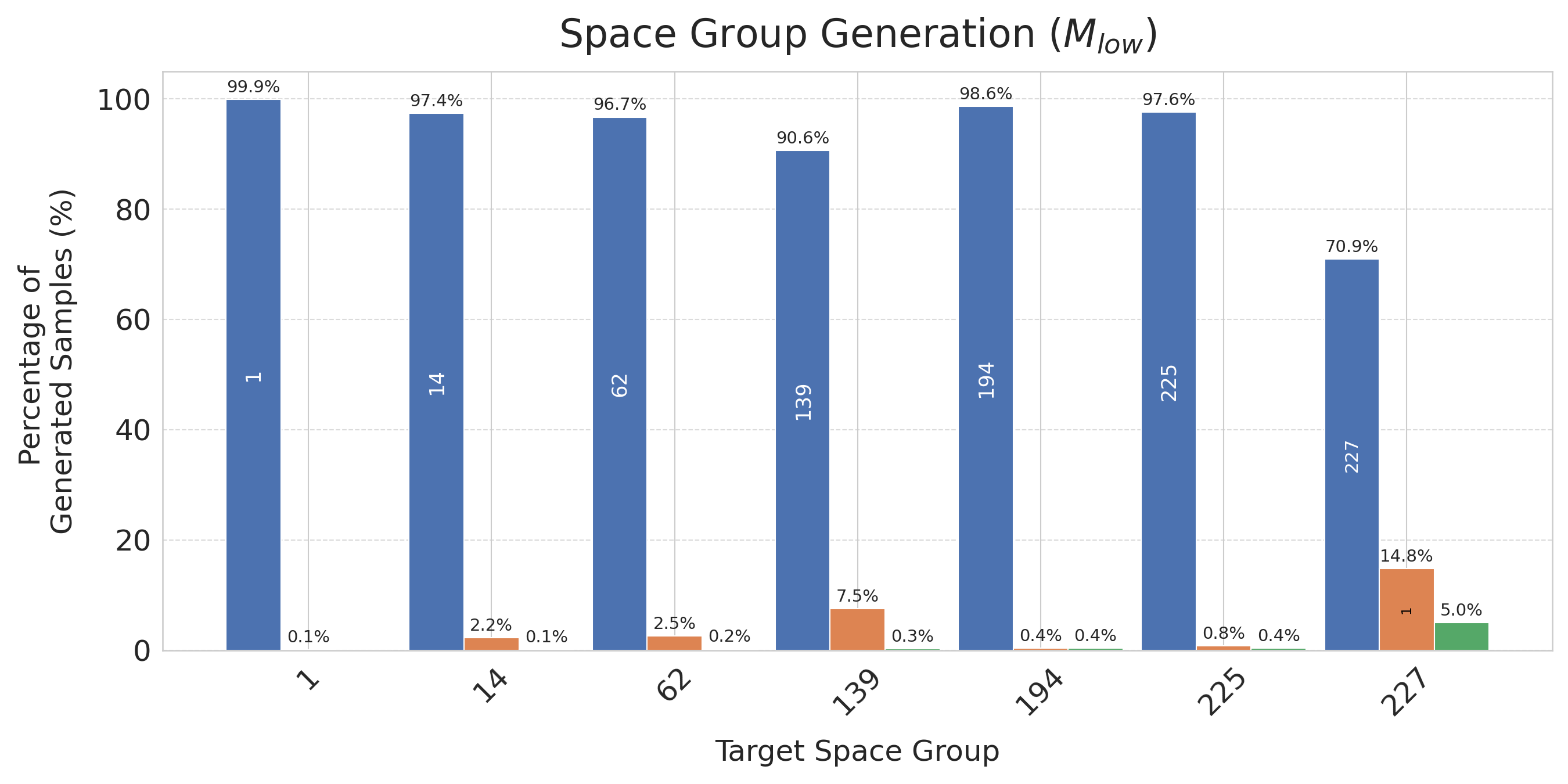}
    \caption{Space group — $M_{low}$}
    \label{fig:sg_small}
  \end{subfigure}\hfill
  \begin{subfigure}{0.48\textwidth}
    \centering
    \includegraphics[width=\linewidth]{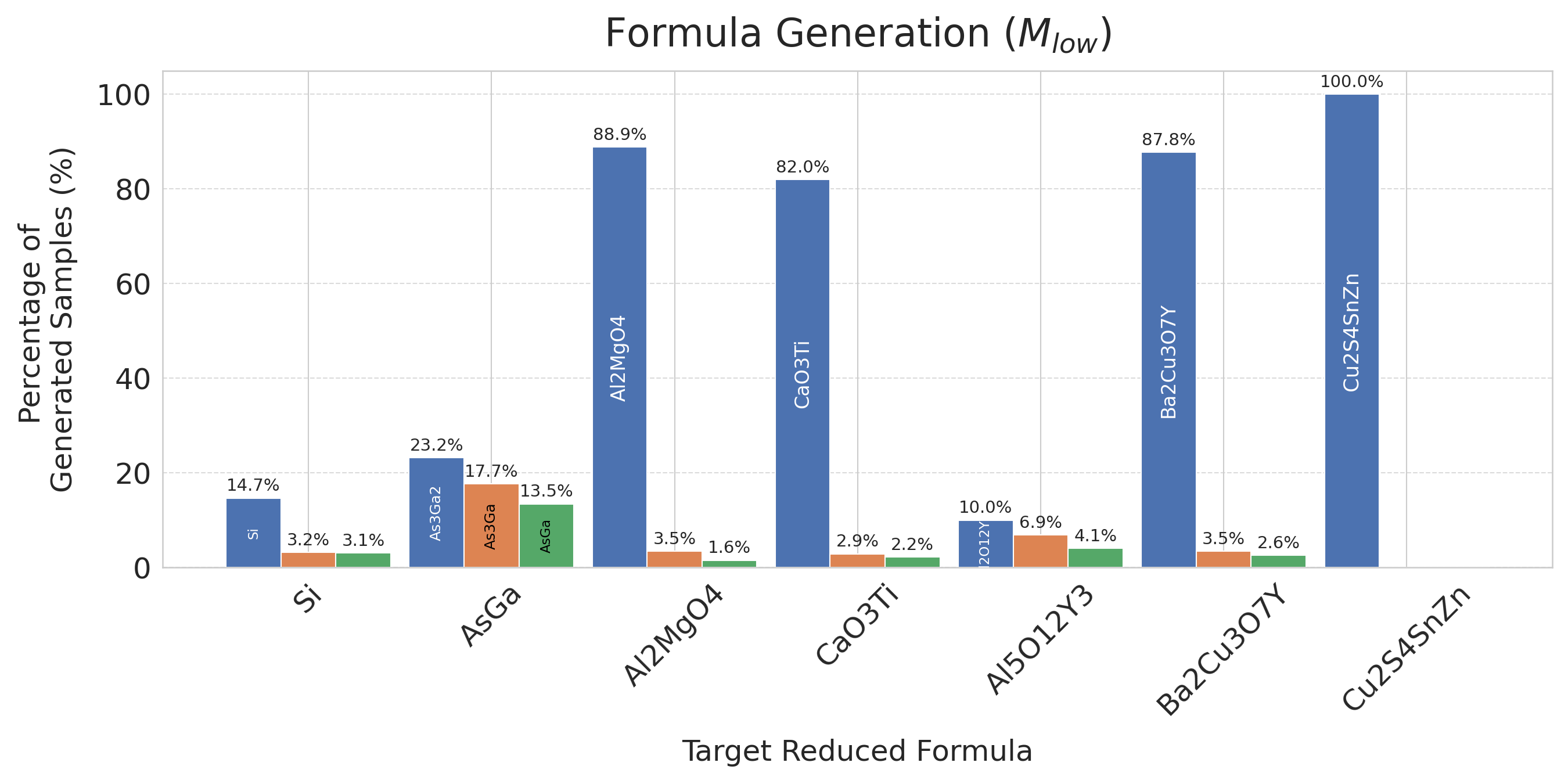}
    \caption{Formula — $M_{low}$}
    \label{fig:formula_small}
  \end{subfigure}

  \vspace{0.6em}

  \begin{subfigure}{0.48\textwidth}
    \centering
    \includegraphics[width=\linewidth]{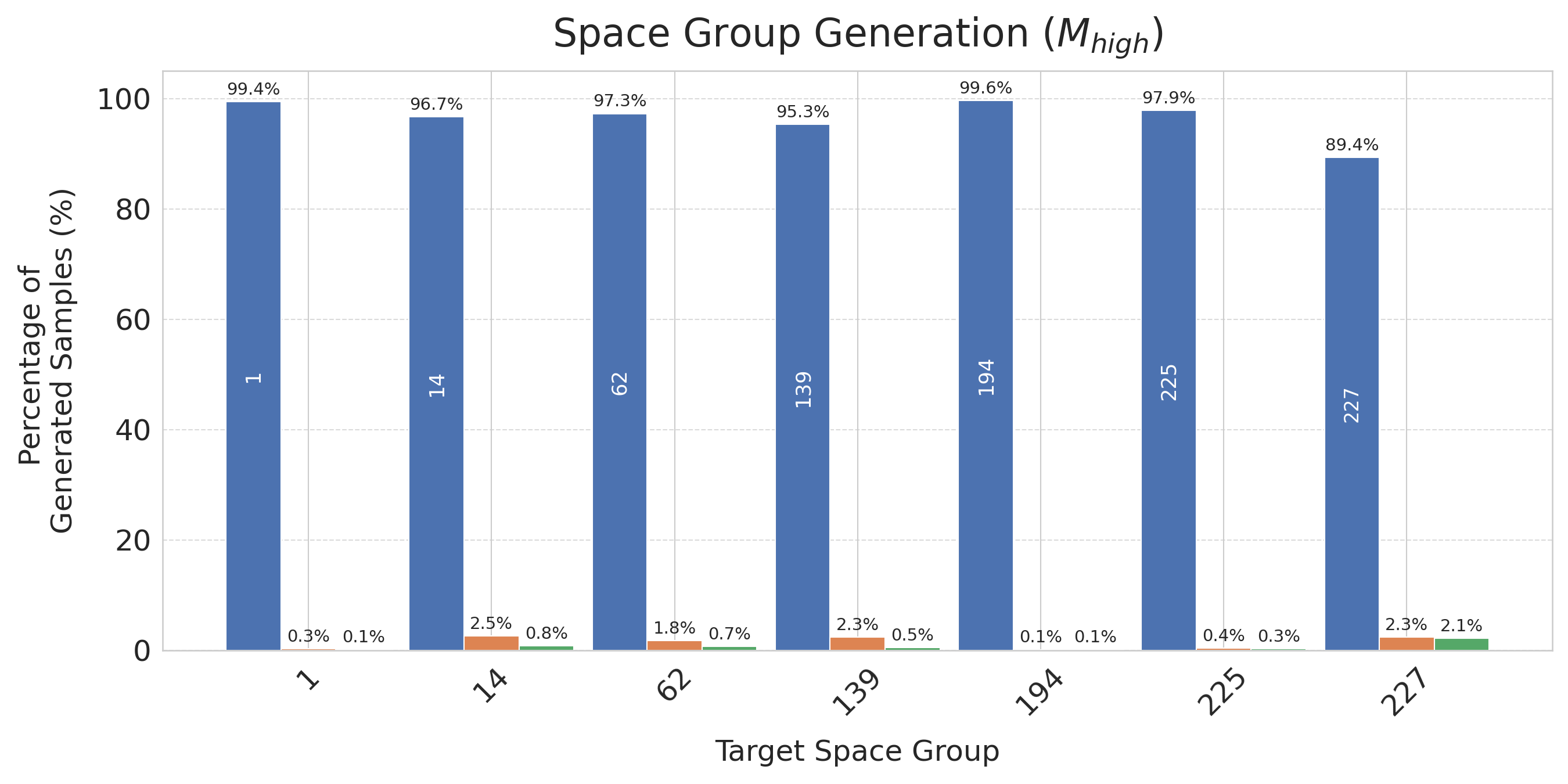}
    \caption{Space group — $M_{high}$}
    \label{fig:sg_large}
  \end{subfigure}\hfill
  \begin{subfigure}{0.48\textwidth}
    \centering
    \includegraphics[width=\linewidth]{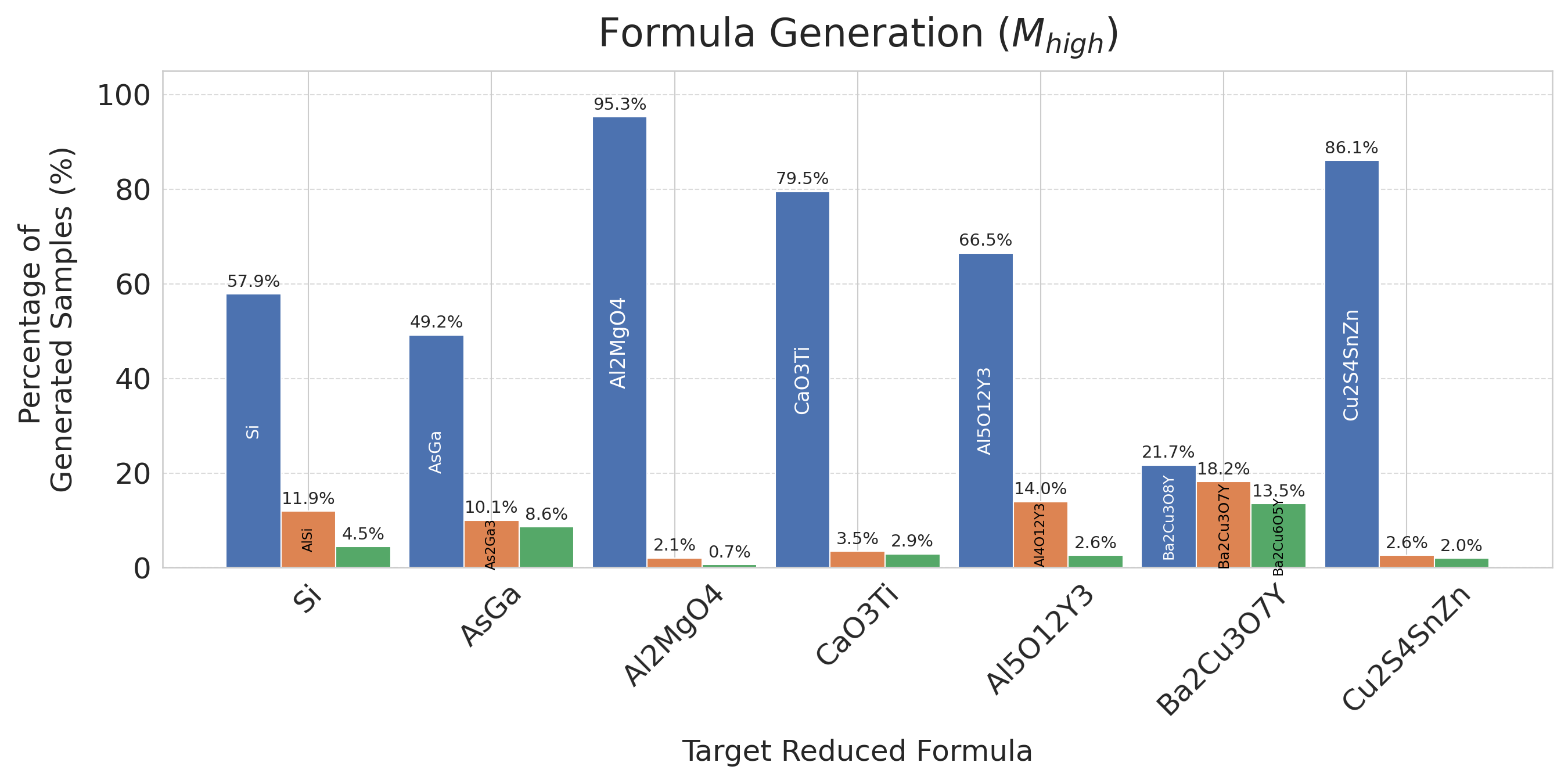}
    \caption{Formula — $M_{high}$}
    \label{fig:formula_large}
  \end{subfigure}

  \caption{Space-group (left column) and reduced-formula (right column) results arranged by model (rows: $M_{xyz}$, $M_{low}$, $M_{high}$). The top 1 guess is blue, the top 2 guess is orange and the top 3 guess is green.}
  \label{fig:sg_formula_3x2}
\end{figure*}
For space group generation, the $M_{high}$ model showed the best overall performance by having almost always the highest percentages for the specified space group. This can be seen in the Figure \ref{fig:sg_large}. The other models where also very good but struggled with the space groups $139$ and $227$, having a lower score there. A similar trend was observed for formula generation, where the $M_{high}$ model demonstrated consistently strong performance. A notable exception was observed for \ce{Ba2Cu3O7Y}, where the model has a significantly lower reproducibility rate than the $M_{low}$ model with a value of $18.2$.

Interestingly, we observed the following differences in the performance across the the various models in this conditional generation task. 
The $M_{low}$ model performed much worse on single and dual elemental compositions, while performing the best on triple and quadruple compositions compared to the other two models. 
The reverse seems to be true for the $M_{xyz}$ model which performs the best, by a small margin, in the single and dual elemental compositions, while performing significantly worse in the triple and quadruple elemental compositions. 
Additionally, $M_{xyz}$ could only reproduce the \ce{Ba2Cu3O7Y} formula right for $7\%$ of the generated materials.
Therefore, the  $M_{high}$ model seems to be again the most consistent model out of the ordering we tested.

\begin{figure*}[t]
  \centering
  \begin{subfigure}{0.32\textwidth}
    \centering
    \includegraphics[width=\linewidth]{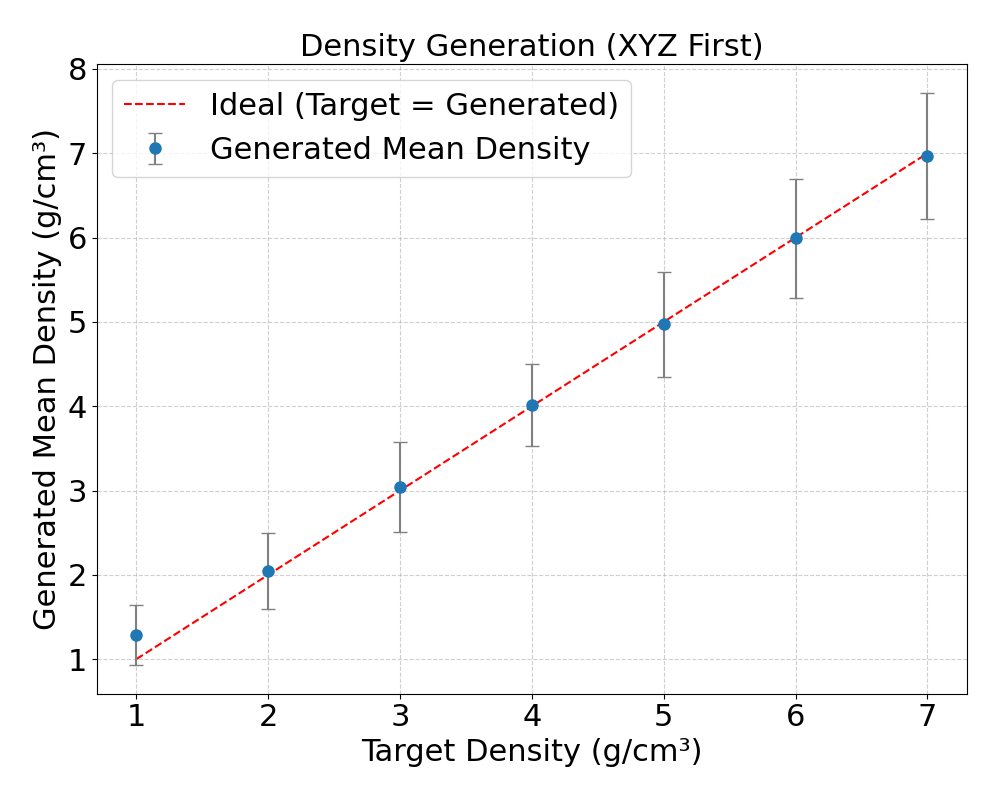}
    \caption{Density plot for the $M_{xyz}$ model.}
    \label{fig:density_xyz} 
  \end{subfigure}\hfill
  \begin{subfigure}{0.32\textwidth}
    \centering
    \includegraphics[width=\linewidth]{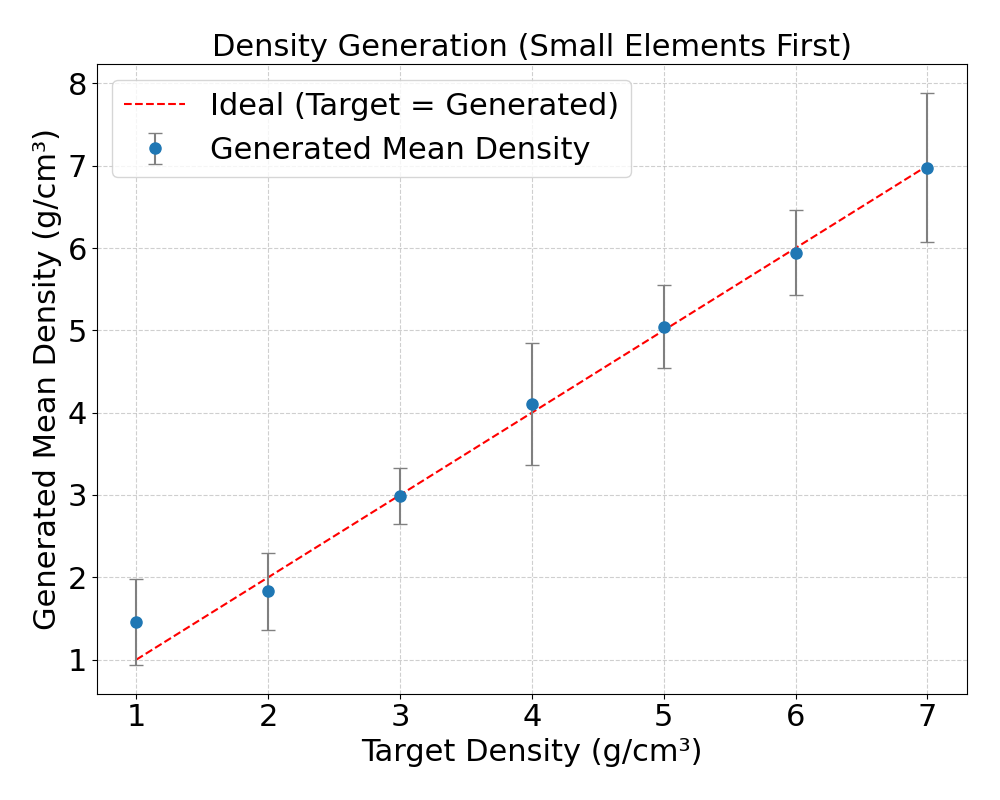}
    \caption{Density plot for the $M_{low}$ model.}
    \label{fig:density_low} 
  \end{subfigure}\hfill
  \begin{subfigure}{0.32\textwidth}
    \centering
    \includegraphics[width=\linewidth]{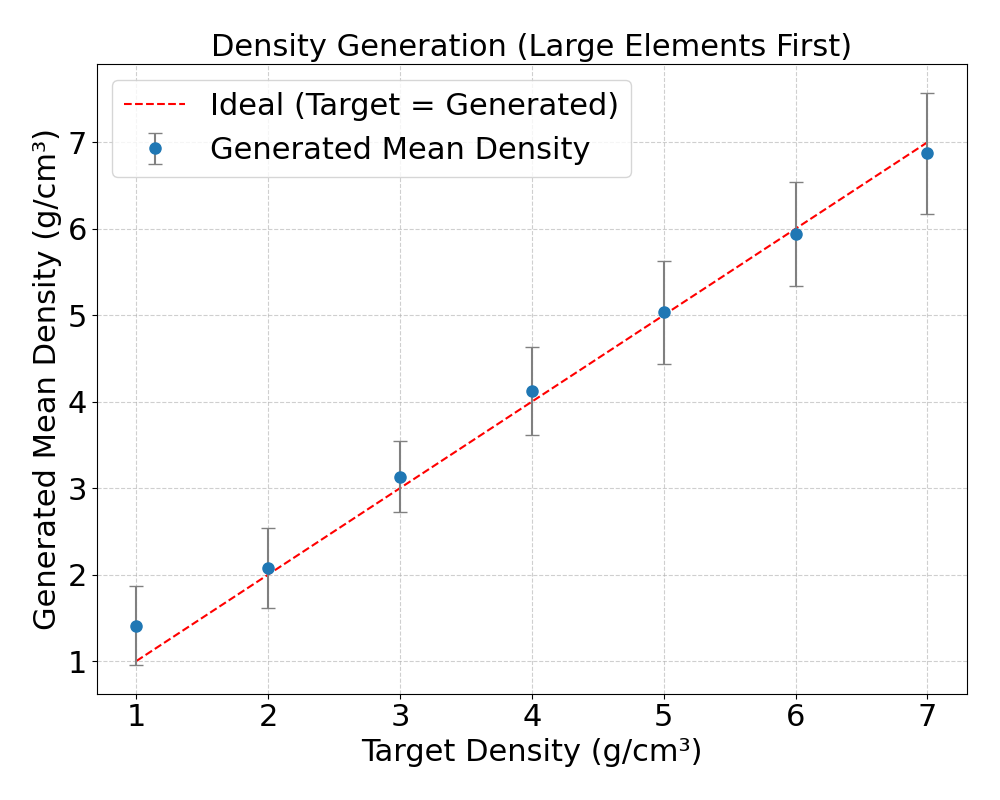}
    \caption{Density plot for the $M_{high}$ model.}
    \label{fig:density_high}
  \end{subfigure}

  \caption{Comparison of the single-conditional densities for the $M_{xyz}$, $M_{low}$, and $M_{high}$ models. The plots illustrate the different predictive distributions learned by each model.}
  \label{fig:density}
\end{figure*}

Lastly, when we tested all three models for the conditional generation on the density of the materials shown in Figure \ref{fig:density}, all models perform very similarly. Each of them can almost perfectly reproduce materials with the required density.

\subsection{Single-conditional Generation of $M_{high}$}
 
After testing the ordering in the previous section, we focused on more advanced single-conditional results for the $M_{high}$ model. Here we tested the model to create materials based on a provided band gap (in eV) and magnetic density (in \si{\cubic\per\angstrom}).
Additionally, we evaluated the models based on their performance on the HHI. The HHI is used to evaluate the scarcity of the generated materials in the US \cite{Gaultois2013}.
We divided the score by $1000$ to bring it in line with the range of scores for the other conditions. Note here also, that all DFT calculations were performed using the Quantum Espresso software package. For all the details for the magnetic density calculation we refer to Appendix \ref{app:A}. For the setup of the band gap calculations we refer to Appendix \ref{app:B}.

\begin{figure*}[t]
 \centering

    \begin{subfigure}[b]{0.475\textwidth}
        \centering
        \makebox[\textwidth]{%
            \includegraphics[width=\textwidth, height=4.5cm]{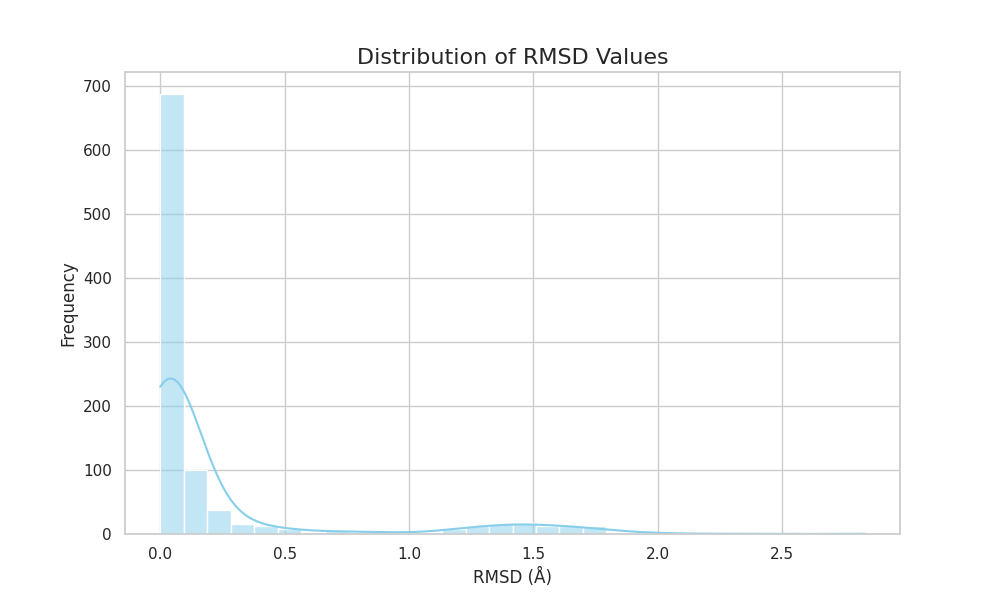}
        }
        \caption[RMSD distribution]%
        {{\small RMSD distribution ($M_{high}$)}}
        \label{fig:rmsd_dist}
    \end{subfigure}
    \hfill
    \begin{subfigure}[b]{0.475\textwidth}
        \centering
        \makebox[\textwidth]{%
            \includegraphics[width=\textwidth, height=4.5cm]{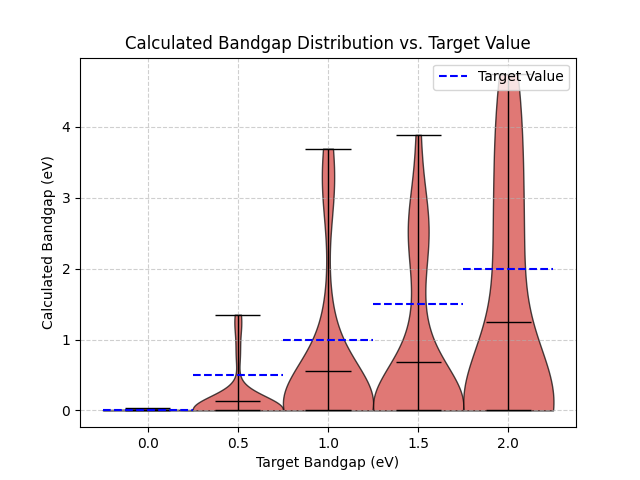}
        }
        \caption[Band gap (DFT)]%
        {{\small Band gap (DFT)}}
        \label{fig:bandgap_violin}
    \end{subfigure}

    \vskip\baselineskip 

    \begin{subfigure}[b]{0.475\textwidth}
        \centering
        \makebox[\textwidth]{%
            \includegraphics[width=\textwidth,height=4.5cm]{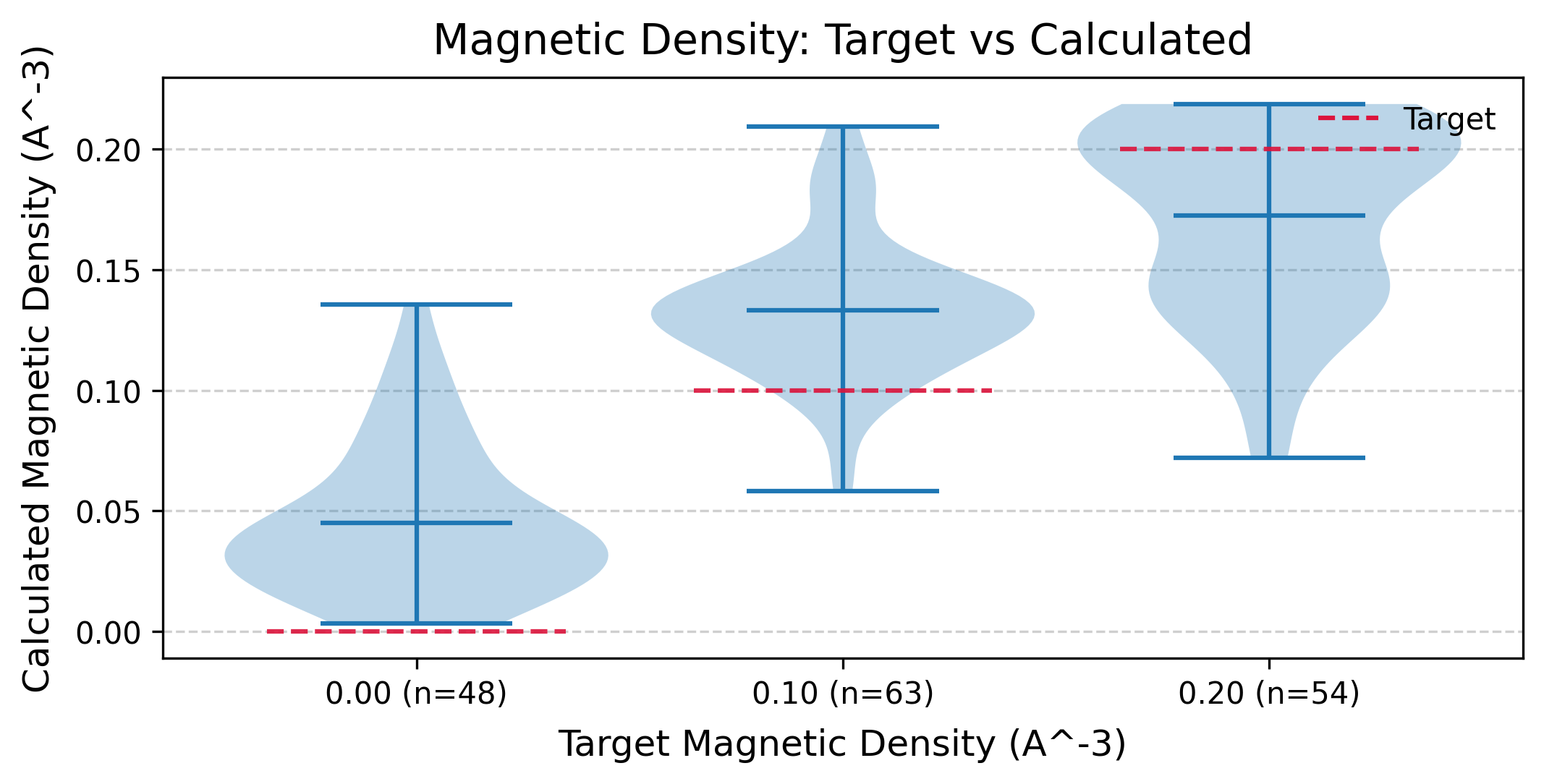}
        }
        \caption[Magnetic density]%
        {{\small Magnetic density}}
        \label{fig:mag_density_violin}
    \end{subfigure}
    \hfill
    \begin{subfigure}[b]{0.475\textwidth}
        \centering
        \makebox[\textwidth]{%
            \includegraphics[width=\textwidth,height=4.5cm, keepaspectratio]{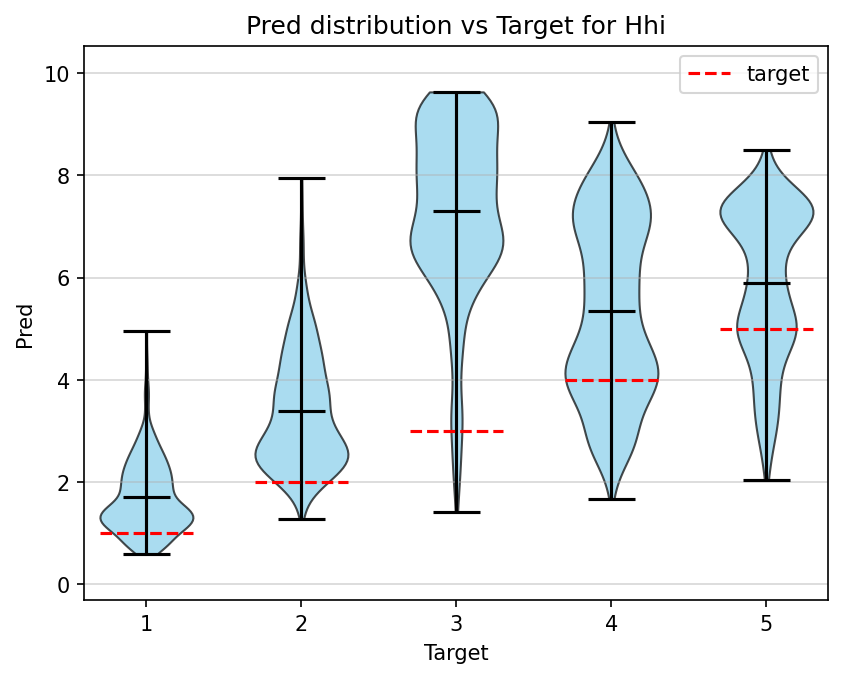}
        }
        \caption[HHI score]%
        {{\small HHI score (divided by $1000$)}}
        \label{fig:hhi_violin}
    \end{subfigure}

  \caption{ RMSD (top-left), band gap (top-right), magnetic density (bottom-left), and HHI score (bottom-right) for $M_{high}$.}
  \label{fig:combined_2x2}
\end{figure*}
As can be seen in the Figure \ref{fig:combined_2x2} in both the band gap and the magnetic density we can see a clear trend that when the requested values get larger, the corresponding generated materials share that property. 
Although this trend emerges, the values still have significant variance in them. 
We assume it is because of the lack of training data in the higher-valued regions (i.e.\ the band gap greater than \SI{1.0}{\eV} and magnetic density greater than \SI{0.1}{\per\cubic\angstrom}) as we have only around $6000$ training samples in these areas.
However, despite the lack of training data the model still generalizes reasonably well.

For the HHI index we can also observe a weak correlation between the generated values and the targets. For smaller values (i.e. $1000$ and $2000$) the model seems to work reasonably well, although it tends to slightly overestimate the values. 
As the target value increases, this trend of overestimation becomes more pronounced, especially for a target of $3000$, where the median of the predicted distribution is significantly higher than the target. For the higher values of $4000$ and $5000$, the model predictions become more aligned with the target values, although the variance in the predictions remains high.

\subsection{Multi Conditional Generation}

We further evaluated the models ability to generate materials based on multi-conditional inputs. Here we only evaluated a small number of conditions based on our time constraints. In this test we generated $512$ samples for each conditions pair. 

\begin{figure*}[t!]
  \centering
  \includegraphics[width=0.48\textwidth]{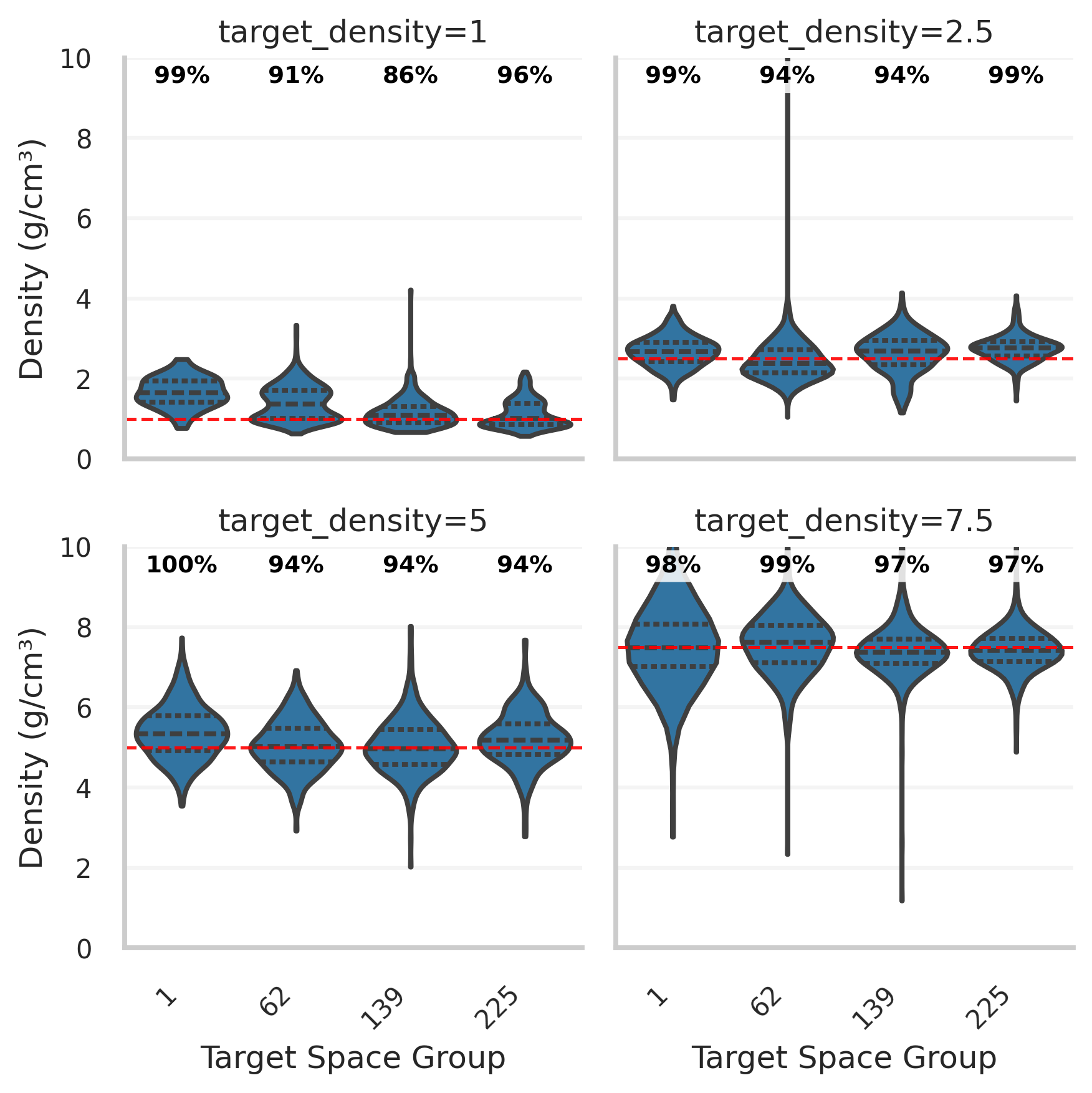}\hfill
  \includegraphics[width=0.48\textwidth]{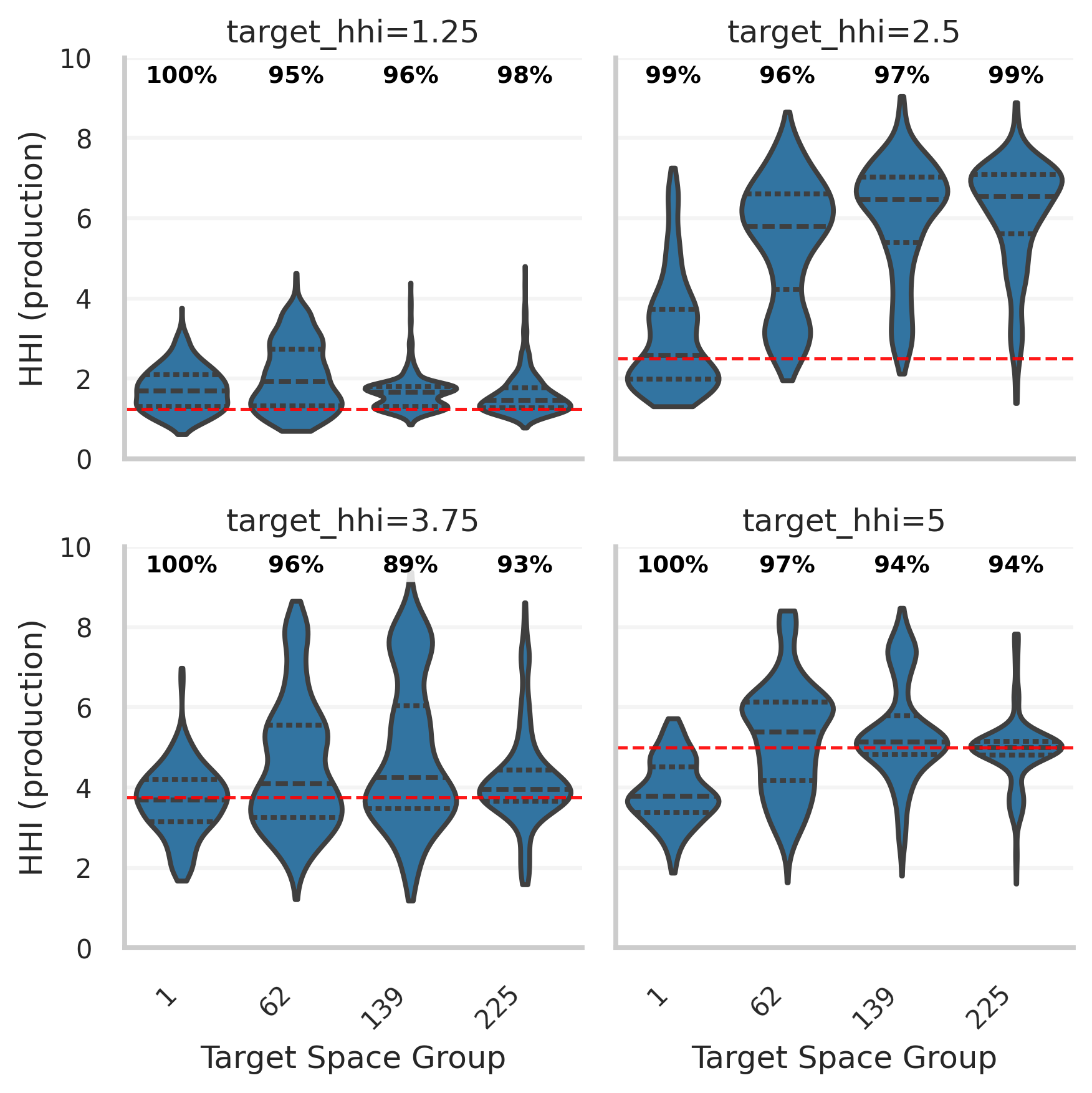}\hfill
  \caption{Multi-conditional generation for the material's density, HHI score (divided by $1000$) and space group}
  \label{fig:multi_cond_density_hhi_sg}
\end{figure*}
First we tested our model on the space group in combination with either the HHI score or the density as can be seen in the Figure \ref{fig:multi_cond_density_hhi_sg}. 
For the materials' density, we observed a strong correlation between the target density and the distribution of generated values across all tested space groups. 
As the target density is increased from \SI{1.0}{\gram\per\cubic\centi\meter} to \SI{7.5}{\gram\per\cubic\centi\meter}, the mean of the generated densities follows this trend. 
This indicates that the model has learned to control the material's density effectively, largely independent of the specified space group. The variance also remains relatively consistent, although it appears slightly larger for higher target densities.
For the HHI score we also observed a slight positive correlation, where the generated material's HHI scores change depending on the target. However, we see here again that the model overestimates the HHI score, especially for lower HHI values, for larger scores the average seems to match better with the target values.

Additionally, the space-group match rate, indicated by the percentages above each graph, was consistently high across conditions, exceeding $90$\% for almost all combinations we tested. This indicates that the model learned to integrate both conditions effectively.

\begin{figure}[h!]
  \centering
  \includegraphics[width=0.8\linewidth]{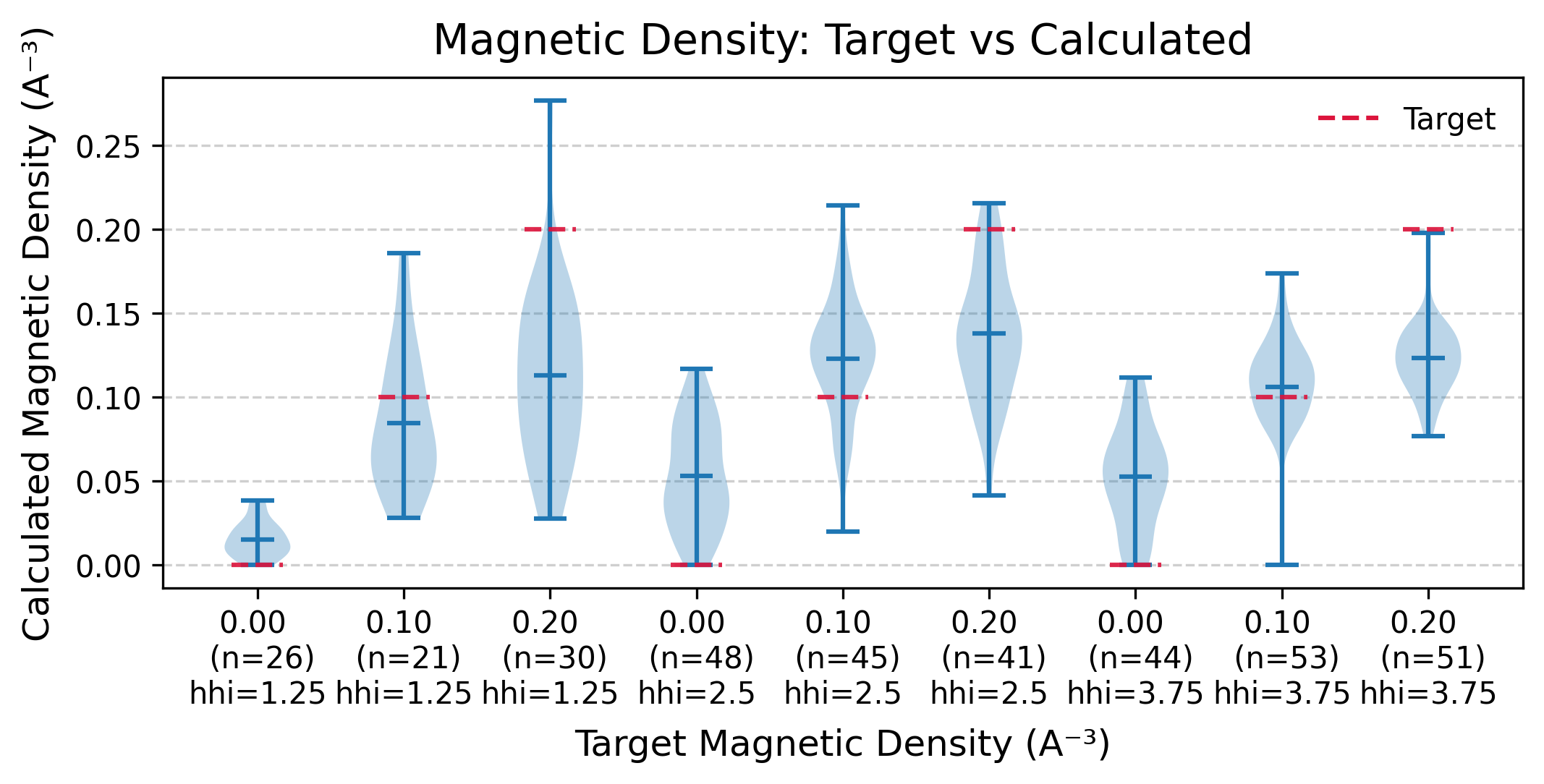}
  \caption{Calculated magnetic density conditioned on both a target magnetic density and a target HHI score (divided by $1000$).}
  \label{fig:mag_density_hhi_violin}
\end{figure}

Furthermore, we evaluated the combination of the magnetic density and the HHI score such that we could generate magnetic materials which are cheap to find.
For the magnetic density, the results in Figure \ref{fig:mag_density_hhi_violin} are very similar to that observed in the single-condition case. 
There is a positive correlation between the target and calculated values, however, like in the single-condition case, the model seems to consistently overshoot the condition at $0.0$ and undershoot it at $0.2$ regardless of the provided HHI score.

\begin{figure*}[h!]
  \centering
  \begin{subfigure}{0.32\textwidth}
    \includegraphics[width=\linewidth]{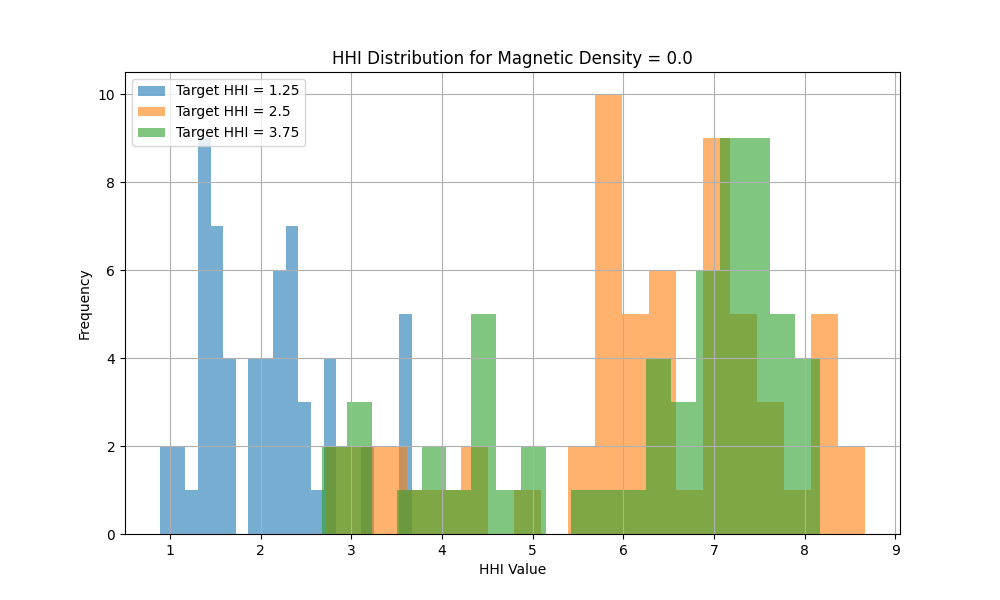}
    \caption{Target Magnetic Density = 0.0}
    \label{fig:hhi_dist_mag_0_0}
  \end{subfigure}\hfill
  \begin{subfigure}{0.32\textwidth}
    \includegraphics[width=\linewidth]{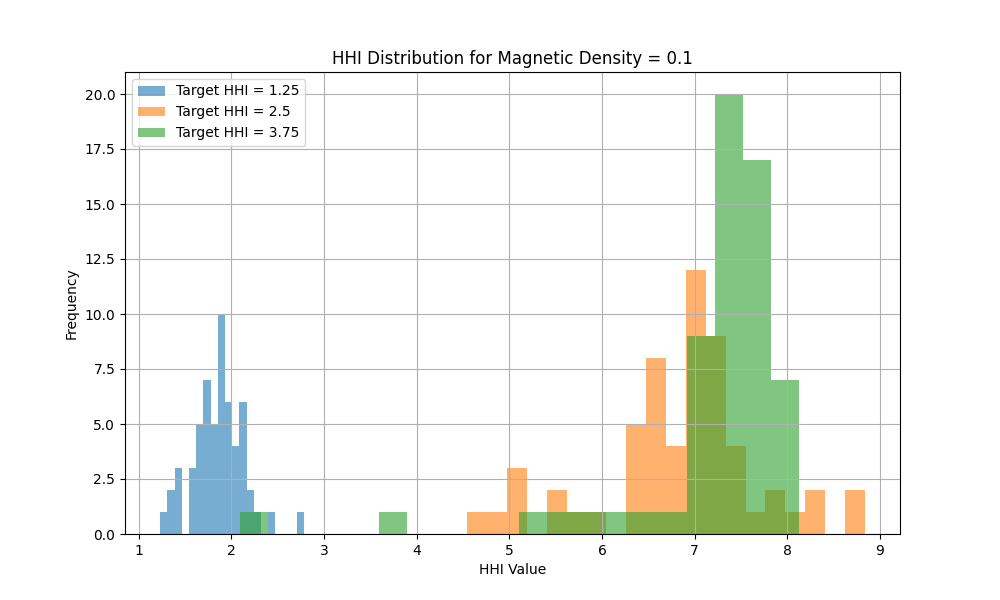}
    \caption{Target Magnetic Density = 0.1}
    \label{fig:hhi_dist_mag_0_1}
  \end{subfigure}\hfill
  \begin{subfigure}{0.32\textwidth}
    \includegraphics[width=\linewidth]{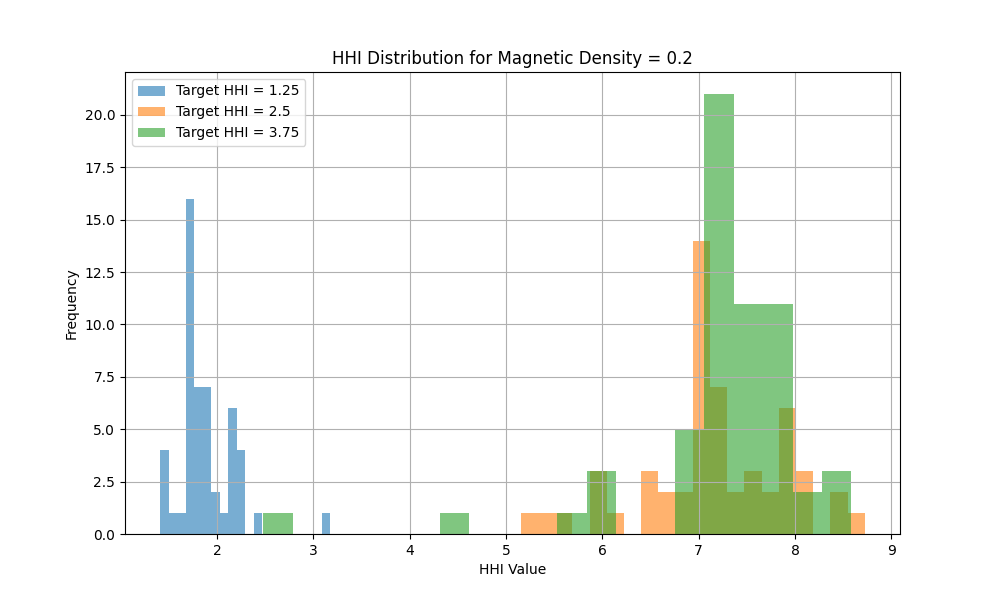}
    \caption{Target Magnetic Density = 0.2}
    \label{fig:hhi_dist_mag_0_2}
  \end{subfigure}
  \caption{Distribution of the generated HHI scores (divided by $1000$) for different target magnetic densities.}
  \label{fig:hhi_dist_by_mag_density}
\end{figure*}

For a high target magnetic density, such as \SI{0.1}{\per\cubic\angstrom} and \SI{0.2}{\per\cubic\angstrom}, c.f.\ Figure \ref{fig:hhi_dist_by_mag_density}, the distribution of the generated HHI values appears to be bimodal. 
This suggests the model generates materials composed of either very common elements (like iron, resulting in a low HHI score) or very uncommon elements (like Gd resulting in a high HHI score), with very few materials generated with HHI values in between. In contrast, for a target magnetic density of \SI{0.0}{\per\cubic\angstrom}, the distribution of HHI is less bimodal, however, there still is a clear bias in the distribution. In general, the model is still not great at reproducing the exact HHI scores, a limitation that was also apparent in the single-condition experiments. This indicates that controlling material scarcity remains a difficult task for the model, especially when trying to achieve specific magnetic properties.

\section{Conclusion}
\label{sec:conclusion}

In this work, we introduced a novel approach for crystal structure generation by representing materials as a sequence of discrete tokens, which can be effectively learned by an autoregressive transformer model.
We tested the model on unconditional and conditional generation tasks, where it achieved strong overall performance. 
However, we observed that the model was less effective at generating novel materials than current state-of-the-art models.

Conversely, a key advantage of our methodology is its computational efficiency. The model requires only approximately four hours of training on a single A100 GPU, a significant reduction compared to competing architectures.
Additionally, there may be ways to fine-tune the model using reinforcement-learning techniques to improve the novelty of the output. 

We believe that our autoregressive approach is a much simpler alternative to diffusion and flow-based models for material generation. By unifying the representation of all crystal structure components into a single sequence, our method circumvents the need for complex, multi-part diffusion processes.
Emerging optimization techniques from the natural language processing field can be readily applied to this model, as it uses the same underlying architecture and training procedure.
In addition, methods such as reinforcement learning, which have shown great promise in language modeling, could also be used to fine-tune our model in terms of novelty or other metrics.
\section*{Acknowledgments}
This work was supported in parts by the Fraunhofer Internal Programs under Grant No. PREPARE 40-08394 (ESPINN).

\section*{Code Availability}
The code will be available after acceptance on \url{https://github.com/Fraunhofer-SCAI/materium}.

\section*{Data Availability}
The data will be available after acceptance on \url{https://github.com/Fraunhofer-SCAI/materium}.

\printbibliography

\begin{appendices}
\section{} \label{app:A}

To evaluate the magnetic properties, we performed spin-polarized (`nspin = 2') calculations for materials containing known magnetic elements (e.g., Fe, Co, Ni, Mn). 
To break the symmetry and guide the self-consistent field (SCF) cycle towards a physically meaningful magnetic ground state, initial magnetic moments were assigned to these atomic species (e.g., 0.8 for transition metals, 0.1 as a default). This setup allows for the determination of the net magnetic density of the relaxed structure.

For all calculations, we employed the Generalized Gradient Approximation (GGA) with the Perdew-Burke-Erzerhof (PBE) functional \cite{PhysRevLett.77.3865} to describe the exchange-correlation effects. The interaction between the ionic cores and valence electrons was modeled using Standard Solid-State Pseudopotentials (SSSP) \cite{sssp}.

The generated structures were fully relaxed using the variable-cell relaxation scheme (`vc-relax'), where both atomic positions and lattice parameters were optimized simultaneously. The structural optimization was considered converged when the total forces on the atoms were below $10^{-3}$ atomic units of force and the total pressure on the unit cell was less than \SI{0.5}{\kilo\bar}. 
The convergence threshold for the self-consistency is $10^{-6}$ Ry and we employed the local Thomas-Fermi mixing scheme with a `mixing\_beta` of 0.4. A kinetic energy cutoff of 60 Ry for the wave functions and 600 Ry for the charge density and potential were used to ensure numerical accuracy. 
Since most systems are metallic, we applied Methfessel-Paxton smearing \cite{Methfessel1989} of 0.01 Ry to the electronic occupations.

\section{} \label{app:B}

For the calculation of the band gap, we mostly used the same relaxation configuration as described in Appendix \ref{app:A}, but the overall procedure is more involved.

For the 'vc-relax' we changed some hyperparameters as we had trouble with the convergence of the generated materials. We reduced the force threshold to $10^{-2}$ and the convergence threshold for the self-consistency to $10^{-4}$ Ry. We increased the cut-off for the plane-wave basis set to $80$ Ry and correspondingly also the kinetic energy cutoff to $800$ Ry. We changed the smearing to Gaussian smearing as Methfessel-Paxton can cause problem in non-metals which we want to generate here.

After the relaxation we first did a self-consistent field (\texttt{calculation = `scf'}) calculation on the fixed, relaxed structure. This step computes the ground-state charge density using a fine, automatically generated Monkhorst-Pack k-point grid to ensure a high level of convergence. The k-point grid for this stage was generated using an automatic scheme targeting a density of 2500 k-points per reciprocal atom (KPPA). The resulting charge density is then used as a fixed input for the subsequent band structure calculation.

Next, a non-self-consistent calculation (\texttt{calculation = `bands'}) is executed. This calculation evaluates the electronic eigenvalues along a high-symmetry path within the first Brillouin zone. The specific high-symmetry points and the path connecting them are determined automatically based on the crystal structure's symmetry group. To ensure a smooth representation of the band dispersion, the path between each high-symmetry point is sampled with 30 k-points.

For the non-self-consistent step, the number of bands (\texttt{nbnd}) is a critical parameter. It was determined by first calculating the total number of valence electrons from the pseudopotential files for each element in the structure. A margin of at least $25$\% additional empty bands was then added to this number. This ensures that a sufficient portion of the conduction band is calculated, allowing for an accurate determination of the band gap.

Finally, the raw output from the 'bands' calculation is processed to collate the energy eigenvalues for each k-point along the high-symmetry path, which can then be plotted to visualize the electronic band structure and extract the band gap.

\section{} \label{app:C}
\begin{figure}[h!]
  \centering
  \includegraphics[width=0.8\linewidth]{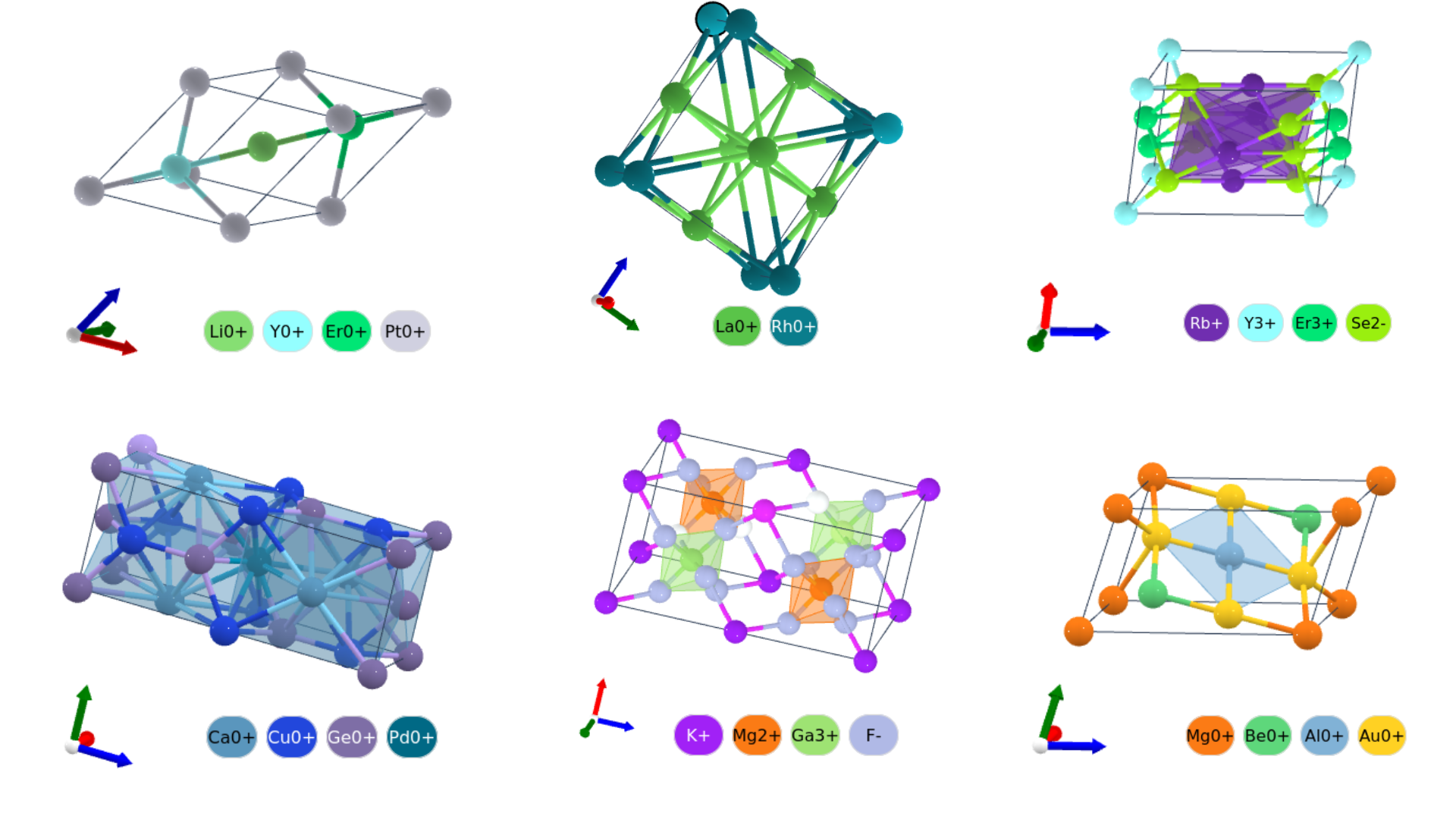}
  \caption{Six unrelaxed samples from the unconditional generation of the $M_{large}$ model visualized.}
  \label{fig:unc_gen_samples}
\end{figure}

\end{appendices}

\end{document}